\documentclass[pdflatex,sn-basic]{sn-jnl}

\usepackage{graphicx}%
\usepackage{multirow}%
\usepackage{amsmath,amssymb,amsfonts}%
\usepackage{amsthm}%
\usepackage{mathrsfs}%
\usepackage[title]{appendix}%
\usepackage{xcolor}%
\usepackage{textcomp}%
\usepackage{manyfoot}%
\usepackage{booktabs}%
\usepackage{algorithm}%
\usepackage{algorithmicx}%
\usepackage{algpseudocode}%
\usepackage{listings}%
\usepackage{subcaption}
\usepackage{gensymb}
\usepackage{hyperref}
\usepackage{bm}
\usepackage{float}
\usepackage{caption}
\usepackage{mathtools}
\usepackage[T1]{fontenc}
\usepackage{makecell}
\usepackage{lineno}

\theoremstyle{thmstyleone}%
\theoremstyle{thmstyletwo}%

\theoremstyle{thmstylethree}%

\begin{document}

\title[Automated Data Engineering and Feature Selection for the Case Study of Warpage Detection in Fused Deposition Modeling]{Automated Data Engineering and Feature Selection for the Case Study of Warpage Detection in Fused Deposition Modeling}


\author[1]{\fnm{Saleh} \sur{Valizadeh Sotubadi}}\email{svalizad@mtu.edu}

\author[1]{\fnm{Nazanin} \sur{Mahjourian}}\email{mahjouri@mtu.edu}

\author*[1]{\fnm{Vinh} \sur{Nguyen}}\email{vinhn@mtu.edu}


\affil[1]{\orgdiv{Department of Mechanical and Aerospace Engineering}, \orgname{Michigan Technological University}, \orgaddress{\street{1400 Townsend Drive}, \city{Houghton}, \postcode{49931}, \state{MI}, \country{USA}}}

\abstract{This study contributes toward development of an Automated Data Processing (ADP) framework designed to evaluate and reinforce optimal machine learning model-feature combinations for predictive tasks in fused deposition modeling (FDM) process datasets. The methodology is centered around a reinforcement learning-inspired policy updating mechanism, where multiple machine learning models are trained on both full feature sets and feature subsets selected through Shapley-based Explainable AI (SHAP XAI) across 217 datasets. At each episode, the framework assesses the predictive accuracy and F1-scores of each model-feature pair, computes a scalar reward, and updates $Q$ values to guide future model selection. SHAP XAI feature importance was employed to generate reduced yet informative feature subsets to enable the framework to explore performance with dimensionality. The policy was shown to evolve over multiple episodes, with reward distributions used to visualize performance stability. Overall, results indicate that leveraging the ADP framework through XAI algorithms successfully converges toward optimal model-feature configurations with improved accuracy and stability. Specifically, the proposed framework improves the test-set AUC from 0.9248 to 0.9731 and increases the mean reward value by more than fifty percent compared with the baseline full-feature configuration.
}

\keywords{Automated Data Processing, Reinforcement Learning, Explainable AI, Shapley, Warpage Detection}



\maketitle

\section{Introduction}\label{sec1}

In recent years, the advent of smart manufacturing through Industry 4.0 has enabled manufacturing processes to benefit from Machine Learning (ML) (\cite{tao2018data}). Several research has been conducted by the manufacturing scholars proving the scalability of ML applications in numerous smart manufacturing processes. Specifically, most research has benefited from data-driven methods through ML models for in-situ monitoring of different manufacturing applications (\cite{ccinar2020machine, bunian2024role, mujtaba2025machine}). Fused Deposition Modeling (FDM) has also benefited from the integration of machine learning methods, particularly for monitoring process stability and identifying print-quality defects such as warpage. Specifically, several studies have explored warpage characterization and detection using data-driven or simulation-based approaches. Researchers in (\cite{moretti2022process}), investigated in-process monitoring of warpage during FDM printing and demonstrated the feasibility of early detection using sensor data, reporting high levels of accuracy and sensitivity. (\cite{syrlybayev2021optimization}), examined warpage through thermomechanical finite element simulations and designed a parameter-optimization framework driven by experimental validation for multi-material prints. Although effective for understanding deformation mechanisms, their approach remained largely simulation oriented and did not incorporate real-time sensor data or ML-based decision strategies. In (\cite{bhandarkar2025warpage}), researchers proposed an image-based deep learning method for detecting warpage in polymer parts and achieved strong classification performance using convolutional neural networks. However, this method relied heavily on large imaging datasets and did not consider resource-efficient or interpretable model development suitable for smaller manufacturing environments. Although ML has been effectively applied to smart manufacturing applications such as warpage detection in FDM, it is evident that developing such frameworks with higher fidelity is difficult to actually execute. Namely, developing an ML framework with higher performance accuracy requires extensive knowledge of the ML domain as well as the manufacturing process itself, and requiring such knowledge is limiting. 

The development of a well-articulated ML framework is not a merely challenging task for manufacturing applications. Generally, several fields of science and engineering encounter significant hurdles in developing ML frameworks capable of achieving higher fidelity. To resolve this issue, Automated Machine Leaning (AutoML) has been researched to bypass the known issues of manually intensive ML development in their respective fields (\cite{Auto_ML_1_2}). Generally, AutoML has been developed for feature extraction, hyperparameter tuning, and ML model selection to automate ML development (\cite{Auto_ML_2_1, Auto_ML_2_2, Auto_ML_2_3}). Similarly, smart manufacturing processes have been benefiting from the notion of AutoML. Prior research has demonstrated an AutoML framework that was designed to identify the best set of hyperparameters for a Support Vector Machine (SVM) model using data gathered for predicting shape errors in milling processes (\cite{denkena2020using}). Similarly, an AutoML solution was provided by applying Enterprise Resource Planning data to resolve the issue of lead time for high-mix parts (\cite{Auto_ML_4_2}). In addition, researchers developed an AutoML framework using four benchmarks to address hyperparameter tuning and model selection through the ML training pipeline to predict the production time during the lithography process (\cite{Auto_ML_4_3}). Using PyCaret (\cite{Auto_ML_4_5}) and AutoKeras (\cite{Auto_ML_4_6}) libraries, prior research has developed two AutoML solutions for classification of ball bearing fault types addressing the hyperparameter tuning and ML model architecture selection (\cite{Auto_ML_4_4}). Furthermore, researchers have developed an AutoML solution using Auto-SKlearn (\cite{feurer-arxiv20a}) to fine-tune the hyperparameters in a SVM model for shape error prediction in milling process where the Mean Square Error (MSE) of the model was effectively reduced (\cite{denkena2020using}). 

While AutoML can aid with automatic model selection and tuning, selecting proper features is still considered an important yet challenging step toward developing a reliable ML framework. Feature selection can make the system faster, cheaper, and more accurate by reducing redundancy and focusing only on the most critical data. As mentioned in prior literature, adding too many sensors can increase costs and lower performance, but focusing on a few key sensors can increase accuracy, lower setup and processing costs, and efficiency (\cite{wang2022solutions}). Traditional ML approaches often rely on manual feature extraction and selection, which are time-consuming and require significant domain expertise to ensure model effectiveness (\cite{wang2018deep}). Studies show that applying ML directly to raw IoT data often results in poor model performance. Careful feature selection is especially critical in manufacturing IoT applications (\cite{shah2020feature}). For example, researchers used a new method to sanitize data signals and then extract important features using time-frequency analysis and adaptive kernel Principal Component Analysis (PCA) to obtain more useful data from audio sensors during milling (\cite{li2019data}). PCA-based methods have also been shown to reduce noise and computation cost in manufacturing fault detection while still capturing key nonlinear and multimodal characteristics of process data (\cite{he2008principal}). In addition, genetic algorithms have been effectively used for selecting and weighting the most relevant features to reduce computational costs and improve model performance, especially in low-resource or real-time environments (\cite{zamalloa2006feature}). To further improve model accuracy in manufacturing, researchers have proposed dynamic feature selection methods that combine genetic algorithms with neural networks. These approaches can outperform traditional techniques such as PCA by better capturing complex patterns in process data (\cite{ghahramani2020ai}). Researchers in (\cite{lei2023mutual}), used mutual information to organize and select the most informative portions of the hourly acceleration data. Using a Residual Attention Network and integrating it with MI, researchers achieved strong performance in detecting abnormal sensor readings and demonstrated good generalization across different bridge datasets. In (\cite{wang2024optimal}), the authors developed a sensor placement algorithm that selected optimal sensor locations by maximizing the MI between measured sensor data and candidate uninstrumented locations. By computing MI from both physical and simulation data, the method identified sensor positions that provided the highest information gain and achieved superior reconstruction accuracy in digital twin applications. Apart from the application of methods like PCA, and MI, Explainable AI (XAI) has been focused as a viable alternative for input feature selection (\cite{wang2025explainable}, \cite{siddique2024explainable}). Unlike PCA and MI, which operate independently of the predictive model and rely solely on statistical or variance-based transformations, XAI identifies important features based on the actual behavior and learned representations of the trained ML model, making it inherently more aligned with model performance and decision quality. Specifically, in (\cite{zacharias2022designing}), XAI was used to derive interpretable feature-selection criteria that guided the choice of the most influential input variables. Using these XAI-selected features, the resulting ML model demonstrated improved predictive accuracy and more stable classification performance compared to training on the full feature set. Additionally, previous research has proved that intuitive input feature selection using XAI improved the ML model performance directed toward in-situ process monitoring in tool wear (\cite{SOTUBADI2025110141}).

As discussed previously, systematic feature engineering plays a crucial role in development of a ML model with higher fidelity in smart manufacturing tasks. In this context, AutoML can significantly benefit from systematic and intelligent feature selection mechanisms to improve model robustness, interpretability, and overall predictive performance. Specifically, a recent study, (\cite{sun2023hybrid}), demonstrated the effectiveness of integrating explainability into automated pipelines for performance enhancement. Similarly, researchers in (\cite{wang2025auto}), introduced an explanation-driven framework for feature selection to improve the trustworthiness of predictive systems. Despite their contributions, these approaches remain limited by fragmented pipeline structures, lack of full integration between feature interpretability and adaptive decision-making, and reduced applicability to complex smart manufacturing environments. Therefore, this paper aims to address the current challenges by contributing toward developing an automated data engineering framework by benefiting from XAI methodologies in smart manufacturing. Hence, the current research contribution seeks to leverage the capabilities of the Automated Data Processing (ADP) framework by enabling intuitive and intelligent input feature selection within the ADP framework. Specifically, the current study aims to benefit from the interpretability capabilities of the developed ML models through XAI approaches. Thus, the extracted feature importance values are utilized to select a subset of input features with higher correlations to the output data. It is expected that removing the redundant input data will help improve the ML model performance. For this purpose, data-driven warpage detection during the in-situ monitoring in Fused Deposition Modeling (FDM) processes is selected as the case study as a proof of concept. Namely, the raw collected data during the FDM process is passed to the ADP framework, which is leveraged with the XAI capabilities, resulting in ML architectures that could predict the occurrence of warpage in the 3D printed parts with high accuracies.

The remainder of this paper is organized as follows. Section~2 describes the experimental setup and the data acquisition procedure used to collect the FDM sensor and process data. Section 3 presents the methodology, including the design of the decision-making unit based on Reinforcement Learning (RL) (Section 3.1), the feature engineering and classification models together with the SHAP-based feature selection strategy (Section 3.2), and the integrated ADP-XAI framework for automated model selection (Section 3.3). Section 4 reports the results and discussion, providing both quantitative and qualitative evaluations of the proposed approach as well as a comparative case study against PCA- and MI-based feature selection. Finally, Section~5 concludes the paper and outlines avenues for future work.

\section{Experimental Setup and Data Acquisition}\label{sec2}

\begin{figure*}[t!]
    \centering
    \includegraphics[width=1.0\linewidth]{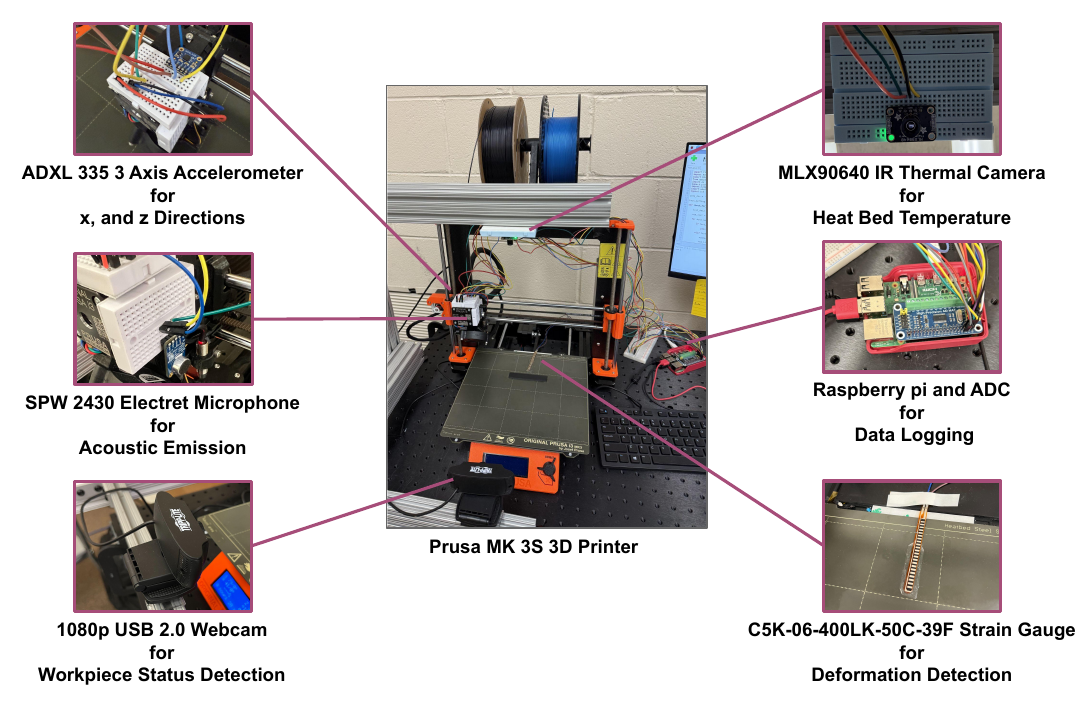}
    \caption{Data collection setup.}
    \label{prelim}
\end{figure*} 

\begin{table}[h]
\centering
\caption{Summary of sensing equipment used in the data collection setup}
\begin{tabular}{lll}
\hline
Sensing Function & Device / Sensor & Model \\
\hline
Vibration Measurement & 3-axis Accelerometer & ADXL 335 \\
Acoustic Emission & Electret Microphone & SPW 2430 \\
Thermal Imaging & IR Thermal Camera & MLX90640 \\
Visual Monitoring & USB Webcam & 1080p USB 2.0 Camera \\
Deformation Detection & Strain Gauge & CSK-06-400LK-50C-3SF \\
Data Logging & Embedded Controller & Raspberry Pi and ADC \\
\hline
\label{tab_equip}
\end{tabular}
\end{table}

Figure \ref{prelim} shows the experimental setup used for data acquisition during the in-situ process monitoring and warpage detection using the FDM 3D printer. To accomplish proper data acquisition during the 3D printing process, a set of sensors were mounted on the printer setup to collect acceleration, acoustic emission, and heat bed deformation data at the sampling rate of 7200 Hz through a High-Precision AD HAT module onto a Raspberry Pi 4.0. Additionally, the temperature distribution of the heat bed was logged using the thermal camera through I2C connection to the Raspberry Pi 4.0. The datasets were collected under different printing conditions to determine whether the printed workpiece was warped or unwarped. Therefore the 1D time series data and the 2D frames of the temperature distributions represented the state of the workpiece for model training and validation, respectively. Table \ref{tab_equip} represents a summary of the sensor modules and data logging equipment that were used in the current study.

Polylactic Acid (PLA) and Acrylonitrile Butadiene Styrene (ABS) materials were used as printing materials to print 3D-printed parts of $L = 80\times W = 20\times H = 5 (mm^3)$ dimension. Table \ref{tab_5_2_1} shows the test configurations during the printing process and data collection. 64 print configurations were implemented to print the PLA parts while 32 printing configurations were utilized for the ABS material. Each configuration was conducted 2 to 4 times to print the workpiece that resulted in a total of 217 datasets. Afterwards, the datasets were preprocessed using a low-pass $FIR$ filter with a cut-off frequency of 3600 Hz ($f_c = 3600 Hz$) and a $ZOH$ technique to reduce background noise and data spikes throughout the time series data. 

Warpage labels were assigned based on visual inspection, where samples exhibiting apparent bending or edge deformation were classified as warped, and this visual assessment approach is consistent with prior studies in FDM warpage analysis (\cite{szalai2025investigation, erokhin2023defects, tanabi2025multi}), where qualitative inspection has been used as a practical means of identifying deformation when precise metrology systems are unavailable or impractical for in-line evaluation. The ratio of warped to unwarped samples was approximately $40\%$ to $60\%$, indicating a mild class imbalance. To mitigate any bias introduced by this skewness, a weighted classification strategy was employed during model training to compensate for the unequal class distribution. In addition, the train and test splits were performed such that all printing parameter configurations were proportionally represented in both sets, ensuring that no specific printing condition dominated either partition and that the reported performance remained representative and reliable.

\begin{table}[]
    \centering
    \small
    \setlength{\tabcolsep}{1.pt}
    \caption{Combination of different print parameters.}
    \begin{tabular}{c|c|c|c|c}
    \hline
        Material Type &  Print Bed Temp & Fill Density & Extruder Height & Extruder Temp \\
        & ($C$) & ($\%$) & ($mm$) & ($C$)\\
        \hline
        PLA & 30, 40, 50, 60 & 20, 50, 80 & 0.1, 0.15, 0.2 & 200, 215, 230 \\
        \hline
        ABS & 70, 80, 100, 110 & 20, 50 & 0.1, 0.15, 0.2 & 225, 250\\
        \hline
    \end{tabular}
    \label{tab_5_2_1}
\end{table}

\section{Materials and Methods}\label{sec3}
\begin{figure*}[t]
    \centering
    \includegraphics[width=1\linewidth]{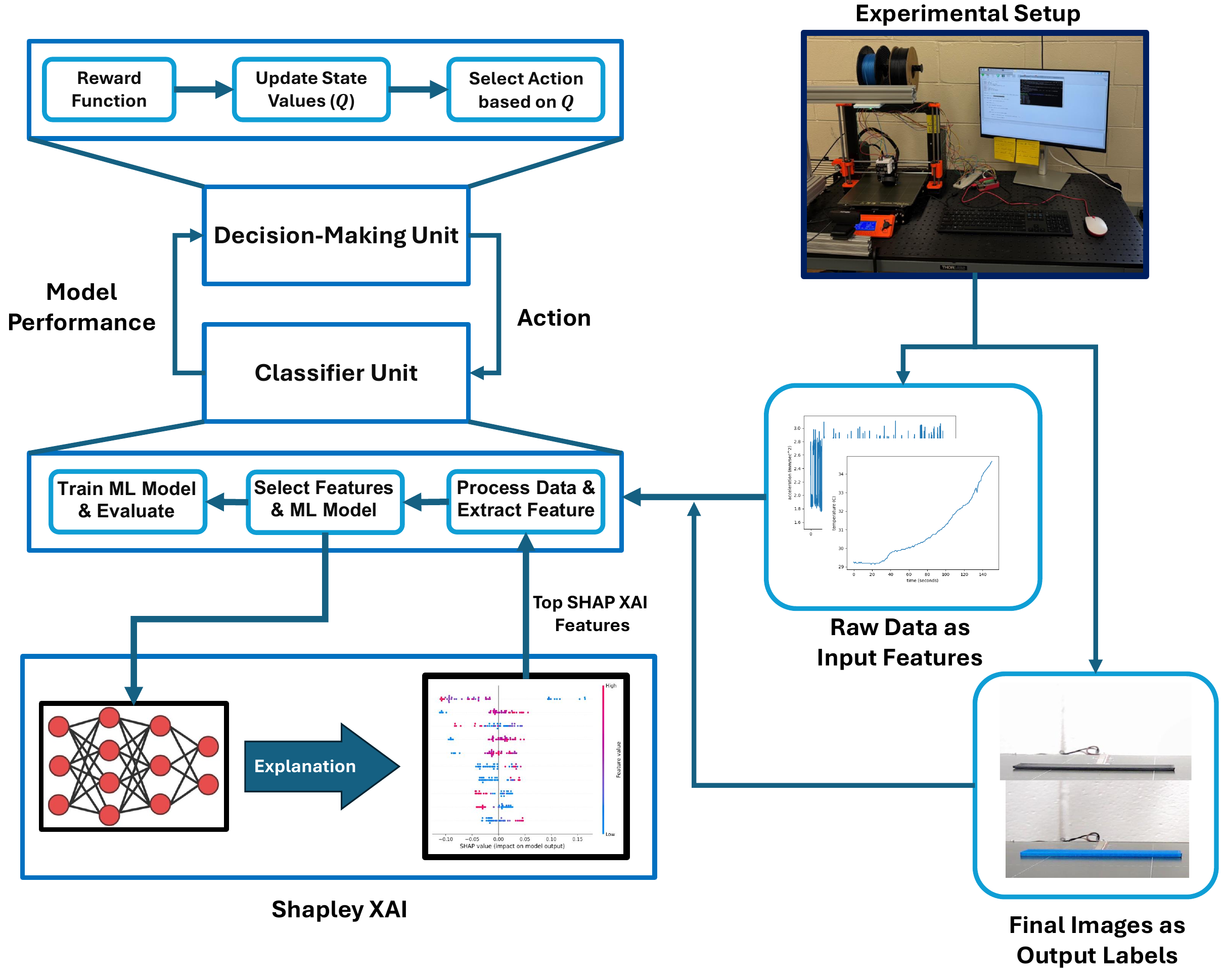}
    \caption{Schematic of the developed ADP framework integrated with XAI for intuitive input feature selection.}
    \label{fig_framework}
\end{figure*}

As shown in Fig. \ref{fig_framework}, the ADP framework consists of a higher-level decision-making unit in conjunction with the lower-level classification unit. Initially, an action is selected by the decision-making unit. The generated action is passed to the lower-level classification unit where the ML framework is developed based on the extracted action within the decision-making unit. Subsequently, the reward function is derived from the quantitative analysis of the ML performance that correlates to the selected action by the decision-making unit. The calculated reward value is inputted into the decision-making unit where the cumulative value of the selected action is extracted. The process continues until the framework converges to an optimal solution defined as the optimal policy. 

\subsection{Decision-Making Unit}

In this work, a Multi Armed Bandit (MAB) was used to represent the decision-making unit. An MAB agent is a fundamental Reinforcement Learning (RL) algorithm that can be framed as a tuple of $(S_i, A_i, R_i)$, where $S_i$ represents the $i^{th}$ state within the search space (\cite{sutton2018reinforcement}). $A_i$ is the corresponding action for $S_i$, and $R_i$ is the reward value extracted through the execution of $A_i$. In the context of the proposed framework, each state corresponds to a specific configuration of feature-extraction and ML-model choices within the action space. The actions represent the selection of a particular ML model and its associated feature subset used for classification. A policy denotes the strategy adopted by the decision-making unit to choose actions based on the evolving Q-values. The reward is the performance feedback obtained from the chosen model, computed using the accuracy, and F1-based reward function. In general, the MAB agent calculates the cumulative reward value of each state namely known as $Q_i$. The value $Q_i$ can be calculated as follows.

\begin{equation}\label{eq_2_1}
    Q_{i}^{m +1} = (1 - \alpha)Q_{i}^m + \alpha R_{i}^m
\end{equation}

\vspace{6pt}

Here, $Q_i^m$ is the $Q$ value of the $i^{th}$ state at the $m^{th}$ iteration. $\alpha$ is the learning rate, and $R_i^m$ is the reward value received by the MAB agent at the $m^{th}$ iteration. 

The action associated to every state is selected based on the $\epsilon$ greedy algorithm. Specifically, the agent selects the action through the exploration-exploitation trade-off. This ensures that the agent is not selecting the action of a sub-optimal state. 

\begin{equation} \label{eq_2_2}
    A \gets \left\{\begin{matrix}
        r \leq \epsilon :  &a & & \gets & randint(a_i|a_i\in A)\\
        &&&&\\
        r >  \epsilon : & a& &  \gets & argmax(Q + c \sqrt{\frac{ln(i)}{Na(a_i)}})
    \end{matrix}\right.
\end{equation}

\vspace{6pt}

Here, $r$ is a random event that is generated at every iteration. In the event that $r$ is less than the $\epsilon$ parameter, the action of a random state is selected by the agent at the given iteration. Otherwise, the selected action is selected through the Upper Confidence Bound (UCB) method. Specifically, in the exploitation scenario, the agent selects the action of a state that has achieved the maximum $Q$ value among other states within the search space.

The learning rate $\alpha$ in the Q-update rule was set to $0.2$. This value provides a balance between incorporating new reward information and maintaining stability in the value updates. Larger $\alpha$ values caused the Q-values to fluctuate excessively during preliminary testing, while smaller values slowed convergence. The exploration parameter $\epsilon$ in the $\epsilon$--greedy strategy was set to $0.1$ to allow the agent occasional exploration and to prevent premature convergence to a suboptimal state. For the UCB-based action-selection strategy, the constant $c$ was set to $0.5$, which controls the strength of the exploration bonus. This moderate value avoids aggressive exploration while still allowing under-sampled states to be evaluated, leading to more balanced learning. 

\subsection{Classifier Unit}

The classifier unit consists of a feature engineering component and a ML model component. Namely, the feature engineering component receives the raw data signals from the experiments and extracts the input features for training the ML model. Subsequently, the ML model component receives the extracted input features from the feature engineering component and trains a binary classification model that outputs a numerical model correlating the input features to the quality of the 3D printed workpiece as the output of the model. Hence, the binary classifier in this work was trained to determine whether a 3D printed workpiece is warped or unwarped. 

\subsubsection{Feature Engineering Component}
As stated, the feature engineering component receives the preprocessed raw signals in the time domain and extracts the relevant input features for training the binary classification model. The input features are denoted as $X = (X_s, X_{m}, X_{p})$. $X_s$ refers to the input features extracted from the raw signals acquired through the sensory modules. All sensor modules were calibrated prior to data collection following the manufacturer's recommended procedures. $X_m$ denotes the printing parameters as the input features. Finally, $X_p$ is the physical feature of the material that is used during the 3D printing process. These variables represent metadata rather than sensor-derived signals, as they are predefined machine and material settings that remain fixed throughout each printing operation. Recall that the sensory modules constitute of extruder acceleration along the x axis and the z axis ($X_s^x$, $X_s^z$), acoustic emission data ($X_s^{a.e}$), the strain gauge data ($X_s^{s.g}$), and the measured plate temperature through the thermal camera ($X_s^{p.t}$). Similarly, the printing parameters include the extruder temperature ($X_m^{e.t}$), extruder height from the surface of the 3D printed workpiece ($X_m^{e.h}$), and the material infill density ($X_m^{i.d}$). Finally, the physical input feature consists of the density of the material ($X_p^{\rho}$). Except for $X_s^{p.t}$, the input features from the time series data of the sensory modules were derived by applying feature extracting techniques on the raw signals within the time domain. Table \ref{tab_feature_engineering} shows the acquired sensory input features within the time domain. The time-domain features extracted from the sensor signals were selected according to well-established signal-processing principles and their relevance in manufacturing process monitoring. Statistical descriptors such as the mean, standard deviation, maximum value, and peak-to-peak amplitude characterize the overall magnitude and variability of the signals. Shape-related metrics including the crest factor, impulse factor, and shape factor capture transient peaks and non-uniform signal behavior that may occur during unstable or imperfect printing conditions. These features are widely used in vibration analysis, thermal monitoring, and sensor-based fault diagnosis. As a result, the selected feature set provides a physically interpretable representation of the information encoded in the FDM sensor data.

\begin{table}[h!]
    \centering
    \caption{Extracted features from the time domain representation of the sensory data.}
    \label{tab_feature_engineering}
    \begin{tabular}{c|c|c}
    \hline
       \multirow{7}{*}{Sensory Data ($X_s$)}  & Mean & ${}^{\mu}\!X_s$\\
         & Standard Deviation & ${}^{\sigma}\!X_s$\\
         & Peak & ${}^{Max}\!X_s$\\
         & Amplitude & ${}^{Amp}\!X_s$\\
         & Shape Factor & ${}^{S.F}\!X_s$\\
         & Impulse Factor & ${}^{I.F}\!X_s$\\
         & Crest Factor & ${}^{C.F}\!X_s$\\
         \hline
    \end{tabular}
    
\end{table}

All features used for model training were normalized through standardization. After the features extracted from the sensor signals were concatenated with the printing-parameter features and the material-property features, each feature was standardized to zero mean and unit variance. Standardization was performed using the parameters computed from the training folds, and the same parameters were applied to the held-out test data. This ensured consistent scaling across all states of the ADP framework and prevented information leakage from the test samples.

\subsubsection{ML Model Component}

The general form of the ML model is represented as ($F: X \rightarrow Y$), where $F$ represents the architecture of the selected ML model that maps the input features $X$ to the desired output of the model which is represented by $Y$. The developed ML model was a non-deterministic discriminative non-linear function that predicts the 3D print quality by predicting the state of the printed part given a set of input features. Specifically, the ML model should determine if a 3D printed was warped or not based on the given inputs. Hence, a binary-classification ML model is required for the defined objective. Due to the complexity of the classification problem and the number of datasets that were collected, a deep learning approach is not guaranteed due to underfitting and poor model performance. Specifically, deep learning models such as two dimensional CNNs were not used in this study for several reasons. The available dataset does not provide the volume necessary to train a reliable deep architecture and previous research in additive manufacturing has shown that deep learning often fails to generalize under limited data conditions. In many manufacturing environments, especially small and medium sized facilities, deep learning systems are also difficult to deploy due to hardware limitations, high computational cost, and reduced interpretability. In this work the thermal image was not intended to be classified directly; it was used only to obtain the temperature distribution of the build plate, which was then converted into engineered features. If two dimensional CNNs were to be combined with the remaining feature based data, the resulting framework would require a multimodal neural network architecture that jointly processes both image and non-image inputs (\cite{SOTUBADI2025110141, valizadeh2025automated}). Although such architectures are feasible, they are considerably more complex and are even more sensitive to data volume and data imbalance. Given the limited dataset available, a multimodal deep learning solution would not be expected to perform reliably. Therefore, the current study benefits from feature-based ML models that are shown in Table \ref{tab_ML_model_ADP}. As shown in Table \ref{tab_ML_model_ADP}, three architectures are used to train the ML model to predict the quality of the 3D printed part. Specifically, each model architecture is assigned an action index that correlates each of the actions of the upper-level decision-making unit to the ML model architecture in the lower-level classifier unit. 

\begin{table*}[h!]
    \centering
    \caption{Selected binary-classification ML models developed within the ADP framework and the correlating sub action index within the action space of the decision-making unit.}
    \label{tab_ML_model_ADP}
    \begin{tabular}{c|c|c|c}
    \hline
    \multicolumn{4}{c}{ML Model Architectures}\\
    \hline
    \makecell{ML model \\sub action ($A_{ML}$)} & 0 & 1 & 2 \\
    \hline
    \makecell{ML \\Architecture} & \makecell{Random \\Forest} & \makecell{XGBoost \\Classifier} & \makecell{Multi Layer\\ Perceptron} \\
    &($RF$) & ($XGB$) & ($MLP$)\\
    \hline
    \end{tabular}  
\end{table*}

\begin{table*}[h!]
    \centering
    \caption{Different input feature extraction strategies and the corresponding sub action index within the action space of the decision-making unit.}
    \label{tab_FE}
    \begin{tabular}{c|c|c|c|c|c|c|c}
        \hline
         & \multicolumn{7}{c}{$X_s$} \\
        \hline
        Sub action index ($A_{FE}$) & ${}^{\mu}\!X_s$ & ${}^{\mathrm{\sigma}}\!X_s$ & ${}^{\mathrm{Max}}\!X_s$ & ${}^{\mathrm{Amp}}\!X_s$ & ${}^{\mathrm{S.F}}\!X_s$ & ${}^{\mathrm{I.F}}\!X_s$ & ${}^{\mathrm{C.F}}\!X_s$ 
         \\
        \hline
        1 & 0 & 1 & 1 & 1 & 1 & 1 & 1  \\
        2 & 1 & 0 & 1 & 1 & 1 & 1 & 1  \\
        3 & 1 & 1 & 0 & 1 & 1 & 1 & 1  \\
        4 & 1 & 1 & 1 & 0 & 1 & 1 & 1  \\
        5 & 1 & 1 & 1 & 1 & 0 & 1 & 1  \\
        6 & 1 & 1 & 1 & 1 & 1 & 0 & 1  \\
        7 & 1 & 1 & 1 & 1 & 1 & 1 & 0  \\
        8 & 1 & 1 & 1 & 1 & 1 & 1 & 1  \\
        \hline
    \end{tabular}
\end{table*}

\subsubsection{Feature Selection}

Prior work has shown that through selecting a subset of input features, the ADP framework could output the optimal policy where the equivalent ML model achieved the most optimal performance (\cite{valizadeh2024integrated}). Therefore, the authors adapted the feature selection strategy from the previous study to search for the optimal ML solution through the ADP. Hence, the action space of the ADP was developed to enable the framework to automatically search through different combinations of input features. Table \ref{tab_FE} shows different configurations of the input feature combinations where the index of each input feature combination correlates to the action space within the ADP framework. Specifically, in Table \ref{tab_FE}, the value 1 at each row denotes that the extracted feature group is selected to feed into the ML model for the training while 0 refers that the specific input feature group is excluded from the input feature set for the ML training and does not contribute to the ML model training. Table \ref{tab_FE} also shows that, at each scenario, one of the feature groupings is excluded from the input feature set that belongs to $X_s$ while the input features from the $X_m$ and $X_p$ sets are consistently included in the feature set. Therefore, this work presents 7 different combinations of input feature configurations. Additionally, one scenario includes all of the input features from the $X_s$ feature set as well as $X_m$, and $X_p$. Hence, this work results in 8 different configurations of input features where the index of each configuration correlates to a specific action space within the decision-making unit. 

As shown in Table \ref{tab_FE}, 8 input combinations were manually selected as initial feature selection strategies. However, given that 33 input features were extracted, manual feature selection by considering all of the input feature combinations leads to larger number of strategies that require substantially extensive action spaces within the higher-level decision-making unit. Note that different combinations of input features could be determined through Equation \ref{eq_combinations}, where $n$ represents the total number of extracted input features, and $k$ represents the number of subset of input features.

\begin{equation} \label{eq_combinations}
    N = \sum_{k=1}^{n} {}^nC_k: \frac{n!}{k!(n - k)!}
\end{equation}

\vspace{6pt}

A large action space within the decision-making units is challenging due to requirement of extensive computation time. Therefore, considerations should be implemented to intuitively select the input feature combinations, which is the main contribution of this research, where an inference-based input feature selection is included in the ADP framework for intuitive and optimal selection of input features for training the ML framework. In this work, Explainable AI (XAI) was leveraged for the selection of input features by extracting the feature contribution affecting the performance of each ML framework.

In this study, a model-agnostic XAI algorithm was used for automatic and intuitive input feature selection. SHapley Additive exPlanation (SHAP) algorithm was chosen as model-agnostic XAI algorithms to extract feature importance for the given ML models of each of the top states. Specifically, the basis for the SHAP algorithm stems from the game theory where the prediction of the model is decomposed among all the input features involved in the decision-making process. This decomposition is achieved by additive feature attribution analysis described as follows (\cite{bennetot2021practical}).

\begin{equation} \label{eq_2_1_b}
    g(x^{'}) = \phi_{0} +  \sum_{i=1}^{M} \phi_{i} x_{i} ^ {'}
\end{equation}

\begin{equation} \label{eq_2_2b}
    \phi_{i} = \sum_{S \subseteq N/i}^{} \frac{|S|!(M - |S| - 1)!}{M!}[F_{X}(S \cup i) - F_{X}(S)] 
\end{equation}

\vspace{6pt}

Here, $g(x^{'})$ is the explanation model, where $x^{'} \in \{0, 1\}^{M}$. $M$ is the number of features, and $\phi_{i} \in R$. Equation \ref{eq_2_2} describes how the SHAP criterion determines the contribution of all the input features in the model predictions. Using this method, the model is trained on all the feature subsets $S \subseteq F$, where $F$ represents the set of all features. To calculate the contribution of each input feature, the model $F_X(S \cup {i})$ is trained by including the feature $i$. Subsequently, model $F_X(S)$ is trained excluding the feature $i$ from the input features. When the two models are trained, the predictions of a specific input $F_{X}$ of the two models are compared together. The differences of the results between two models are the indications of each input feature's effect on the overall decision of the ML model.

\subsection{ADP Framework}

\begin{figure*}[t!]
    \centering
    \includegraphics[width=1\linewidth]{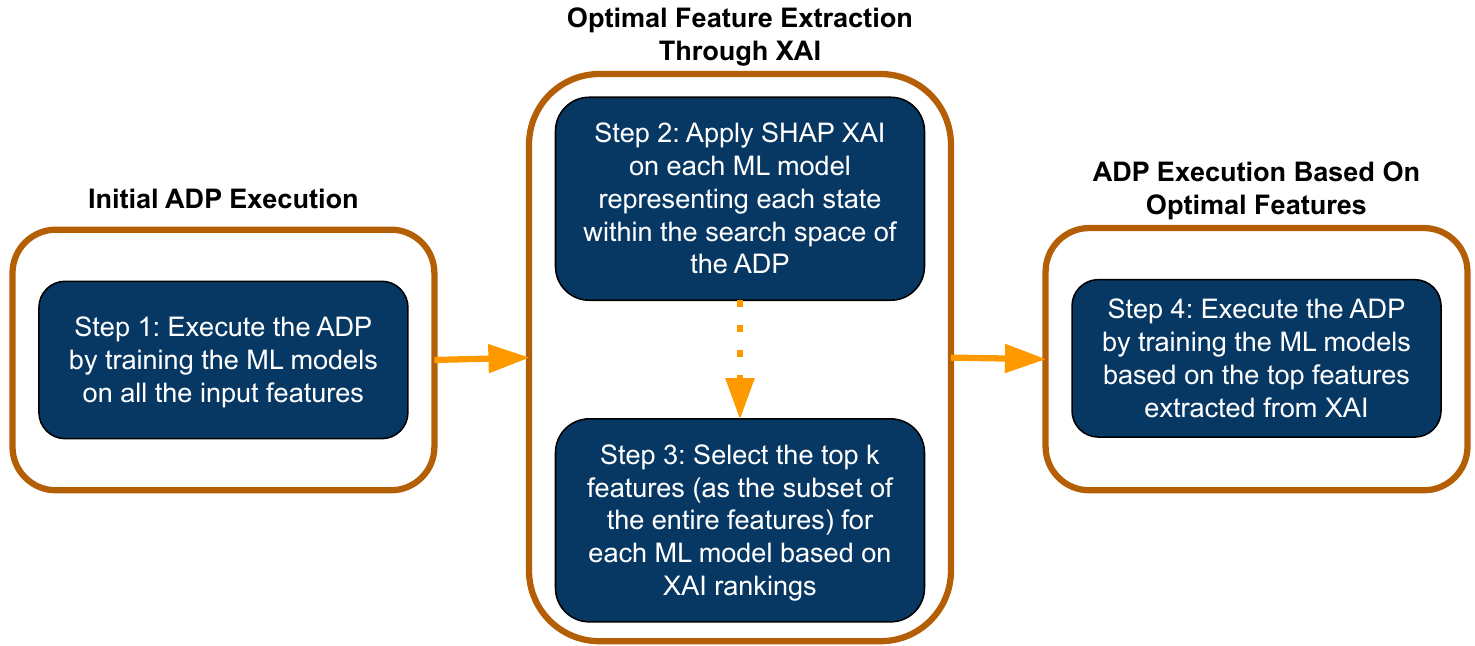}
    \caption{An illustration of the procedural flow of the developed ADP framework for optimal ML model selection.}
    \label{fig_ADP_Hierarchy}
\end{figure*}

Figure \ref{fig_ADP_Hierarchy} describes the automated data selection in detail. The detailed flowchart illustrating the operational structure and execution sequence of the ADP framework has been presented in prior work (\cite{valizadeh2025automated}), where each stage of the decision making and policy update process is comprehensively described. Initially, the ADP is executed to search for the optimal policy that binds the actions of the state space of the decision-making unit to a specific ML framework within the classification unit. Therefore, the general procedure of the ADP framework was developed to solve an optimization problem. Hence, a fit function was required to constraint the optimization problem and satisfy the objectives of the ADP. The developed fit function can be formulated as shown in Equation \ref{eq_bga_reward}, where $Rew_{acc}$ and $Rew_{F1}$ are the nonlinear representations of the ML model accuracy and $F1$ score outputted through the sigmoid function. The final fit function score is the weighted combination of the two terms as follows. To clarify the interpretation of the reward and Q-values, the reward function in Eq. \ref{eq_bga_reward}, produces values that lie approximately within the interval $[-0.5,\,0.5]$ due to the sigmoid transformation and the centering at $0.5$. Consequently, the Q-values do not represent raw classification metrics such as accuracy or F1 score but instead a smoothed estimate of the expected reward. The Q-values evolve according to the incremental update rule, which is represented in Eq. \ref{eq_2_1} with a learning rate $\alpha = 0.1$, which stabilizes learning and prevents abrupt fluctuations. Thus, numerical changes however marginal correspond to meaningful improvements within this bounded reward space and should not be interpreted on the same scale as the original performance metrics.

\begin{equation} \label{eq_bga_reward}
    \left\{\begin{matrix}
        Rew_{acc} = \frac{1}{1 + exp(-(acc - 0.9))} - 0.5\\
        &&&&\\
        Rew_{F1} = \frac{1}{1 + exp(-(F1 - 0.9))} - 0.5\\
        &&&&\\
        R = w_1 \times Rew_{acc} + w_2 \times Rew_{F1}\\
    \end{matrix}\right.
\end{equation}

\vspace{6pt}

The reward function in Equation~\ref{eq_bga_reward} is the weighted sum of two complementary performance terms and is not arbitrarily defined. The first term applies a sigmoid transformation to the model accuracy with a center value of $0.9$. This threshold reflects the target performance level required for FDM warpage-detection classifiers: accuracies below $90\%$ receive a penalizing reward, whereas accuracies above this target provide positive reinforcement. The sigmoid maps the accuracy into a bounded $[0,1]$ interval with a mid-point at $0.5$, which stabilizes the Q-value updates and prevents abrupt changes in reward. The second term incorporates the F1-score, which provides additional robustness for datasets with class imbalance. Because the F1-score jointly captures precision and recall, it evaluates the model's ability to correctly identify minority-class defects which is an essential requirement for manufacturing-quality prediction. By combining accuracy and F1, the reward encourages the ADP agent to select configurations that achieve both high overall accuracy and balanced minority-class detection, resulting in more stable and reliable convergence of the learned decision policy.

Additionally, the weights $w_1$ and $w_2$ were both set to $0.5$ so that the accuracy-based and F1-score--based terms contribute equally to the final reward to ensure that both overall correctness and balanced performance under class imbalance are considered.

The accuracy and the $F1$ score of any ML model can be extracted from Equations \ref{eq_3_1_1} and \ref{eq_3_1_5}, respectively, presented as follows. Here, $TP$, $TN$, $FP$, and $FN$ are the elements of the confusion matrix that represent quantitative measures that compares the ML prediction with the true labels for the given set of instances. Table \ref{tab_3_0} shows the different ML output and true label comparisons in the current study. 

\begin{equation} \label{eq_3_1_1}
    acc = \frac{TP + TN}{TP + TN + FP + FN} 
\end{equation}

\begin{equation} \label{eq_3_1_5}
    F1 = \frac{TP}{TP + 0.5(FP + FN)}
\end{equation}

\vspace{6pt}

\begin{table}[h]
    \centering
    \caption{Different cases comparing the ML model output and the actual labels for the given 3D printed workpieces.}
    \begin{tabular}{c|c|c}
    \hline
         \multicolumn{3}{c}{Variables of the Confusion Matrix}  \\
         \hline
         & Real Label & Model Prediction \\
         \hline
         True Positive (TP) &  Warped & Warped \\
         True Negative (TN) & Unwarped & Unwarped \\
         False Positive (FP) & Unwarped & Warped \\
         False Negative (FN) & Warped & Unwarped\\
         \hline
    \end{tabular}
    \label{tab_3_0}
\end{table}

The reward value (Eq. \ref{eq_bga_reward}) extracted from accuracy and the $F1$-score of each ML framework is subsequently utilized to update the $Q$ value of the corresponding state within the state space of the decision-making unit. Specifically, the ADP should be executed for multiple iterations since the decision-making unit is principally a RL algorithm. For more clarity it is worth mentioning that within the proposed ADP framework, a \emph{state} represents a specific combination of feature-extraction choices and ML model configurations defined in the action space. The \emph{actions} correspond to selecting one of these configurations for training and evaluation. A \emph{policy} denotes the strategy used by the decision-making unit to choose actions based on the evolving Q-values, and the \emph{reward} is the performance feedback returned by the selected model using the accuracy- and F1-based reward function.

As seen in Fig. \ref{fig_ADP_Hierarchy}, in the initial execution of the ADP framework, all ML models corresponding to each state are trained using the full set of input features (Step 1 in Fig. \ref{fig_ADP_Hierarchy}). Subsequently, XAI is employed on each of the ML models to extract the feature rankings of the ML model. SHAP XAI is applied on each model architecture for 10 episodes consecutively where at each iteration the training and testing order of the datasets are shuffled and the ML model is trained entirely on the newly shuffled datasets. Through this process, the feature importance of the input features are extracted at each episode by deriving the mean XAI values for the input features (Step 2 in Fig. \ref{fig_ADP_Hierarchy}). Thereafter, the top $k$ features based on the XAI ranking are selected for each of the ML model architectures that represent the optimal subset of the entire input feature space resulting in high model performances (Step 3 in Fig. \ref{fig_ADP_Hierarchy}). Finally, the ADP is executed benefiting from the optimal subset of the input features that are extracted through the XAI analysis (Step 4 in Fig. \ref{fig_ADP_Hierarchy}). 

\begin{algorithm}[t]
\small
\caption{State-Wise SHAP-based Feature Ranking and Optimal Feature Count Selection.}
\label{algo_shap_optimal_k}
\begin{algorithmic}[0]  
\Statex \textbf{Input:} States $S = \{s_0, \dots, s_{23}\}$, Dataset $D$
\Statex \textbf{Input:} Update rate $\eta$
\Statex \textbf{Init:} $FI_{\text{avg}}[s_i] \gets 0$, $K_{\text{opt}}[s_i] \gets 0$ for all $s_i$
\For{$i \gets 0$ \textbf{to} $23$}
    \State Init ML model pipeline $M_i$ for $s_i$
    \State $FI_{\text{sum}} \gets 0$
    \For{$e \gets 1$ \textbf{to} $10$}
        \State Shuffle $D[s_i]$, split into $D_{\text{train}}, D_{\text{test}}$
        \State Train $M_i$ on $D_{\text{train}}$
        \State Fit SHAP explainer on $D_{\text{train}}$
        \State $FI \gets$ SHAP($M_i$, $D_{\text{test}}$)
        \State $FI_{\text{sum}} \gets (1 - \eta)FI_{\text{sum}} + (\eta) FI$
    \EndFor
    \EndFor

\Statex

\For{$i \gets 0$ \textbf{to} $23$}
    \State Sort $FI_{\text{avg}}[s_i]$ descending $\Rightarrow$ ranked features
    \State $A_{\max} \gets 0$
    \For{$k \in \{5, 10, 15, 20, 25, \text{All}\}$}
        \State Select top-$k$ features $\mathcal{F}_k$ from $FI_{\text{avg}}[s_i]$
        \State Shuffle $D[s_i]$, split into $D_{\text{train}}, D_{\text{test}}$ with $\mathcal{F}_k$
        \State Train new model $M_i^k$ on $D_{\text{train}}$
        \State $A_k \gets$ accuracy of $M_i^k$ on $D_{\text{test}}$
        \If{$A_k > A_{\max}$}
            \State $A_{\max} \gets A_k$
            \State $K_{\text{opt}}[s_i] \gets k$
        \EndIf
    \EndFor
\EndFor
\Statex \textbf{Output:} $K_{\text{opt}}[s_i]$ for each $s_i$
\end{algorithmic}
\end{algorithm}

Algorithm \ref{algo_shap_optimal_k} shows the optimal input feature selection for each ML model based on SHAP XAI rankings. Specifically, each ML model within the lower-level classifier unit associated with every state $s_i$ in the higher-level decision-making unit has an array $FI_{avg} \subset R^{1 \times 24}$, where $i \in \{0, 1, ..., 23\}$ indicates the index of the state $s$ within the search space. As represented in Algorithm ~\ref{algo_shap_optimal_k}, the ML model $M_i$ associated with $s_i$ is iteratively trained over 10 episodes using randomly shuffled subsets of the corresponding dataset. SHAP XAI is subsequently applied to the test set to estimate the contribution of each input feature. The resulting feature importance scores are aggregated using an exponential moving average governed by the parameter $\eta$ to produce a smoothed estimate of the feature relevance profile for each state. To determine the optimal number of top $k$ features for each state, the corresponding ML model for each state is executed consecutively for several iterations, where for every iteration $i$ the hyperparameter $k$ is selected such that $\{k_i \mid k_i \in \{5, 10, 15, 20, 25, \text{All}\}\}$. For each value of $k_i$, the top-$k_i$ features are selected based on the previously computed SHAP-based feature importance scores. The model is then retrained and evaluated using only the selected features, and the resulting prediction accuracy is recorded. The value of $k_i$ that yields the highest accuracy is identified as the optimal number of input features for the corresponding state and is stored accordingly.

\section{Results and Discussion}\label{sec4}
The ADP framework was executed using an Intel Core i5-10600 processor, 16 GB of RAM local machine. As expressed previously, 217 datasets were generated during the FDM process. 80$\%$ of the datasets were used for the training while the remaining datasets were used as test data for evaluation. For each printing-parameter combination, exactly 80\% of its samples
were assigned to the train/validation pool and the remaining 20\% were held out exclusively
for testing. This grouped assignment prevents leakage by ensuring that both subsets contain
representative samples from every condition, while avoiding situations in which all samples
from a given condition fall into only one subset. The 80\% train/validation portion was then evaluated using a 5-fold cross-validation procedure. After every policy determined the ML model and the training framework, the associated ML model would be trained for 10 episodes, and the model was evaluated after each training process to provide the means to calculate the average accuracy and the $F1$ score. These criteria would later be implemented to calculate the reward function to update the $Q$ value of the state corresponding to the action through which the ML framework was developed.

\begin{figure*}[h!]
    \centering
    \includegraphics[width=1\linewidth]{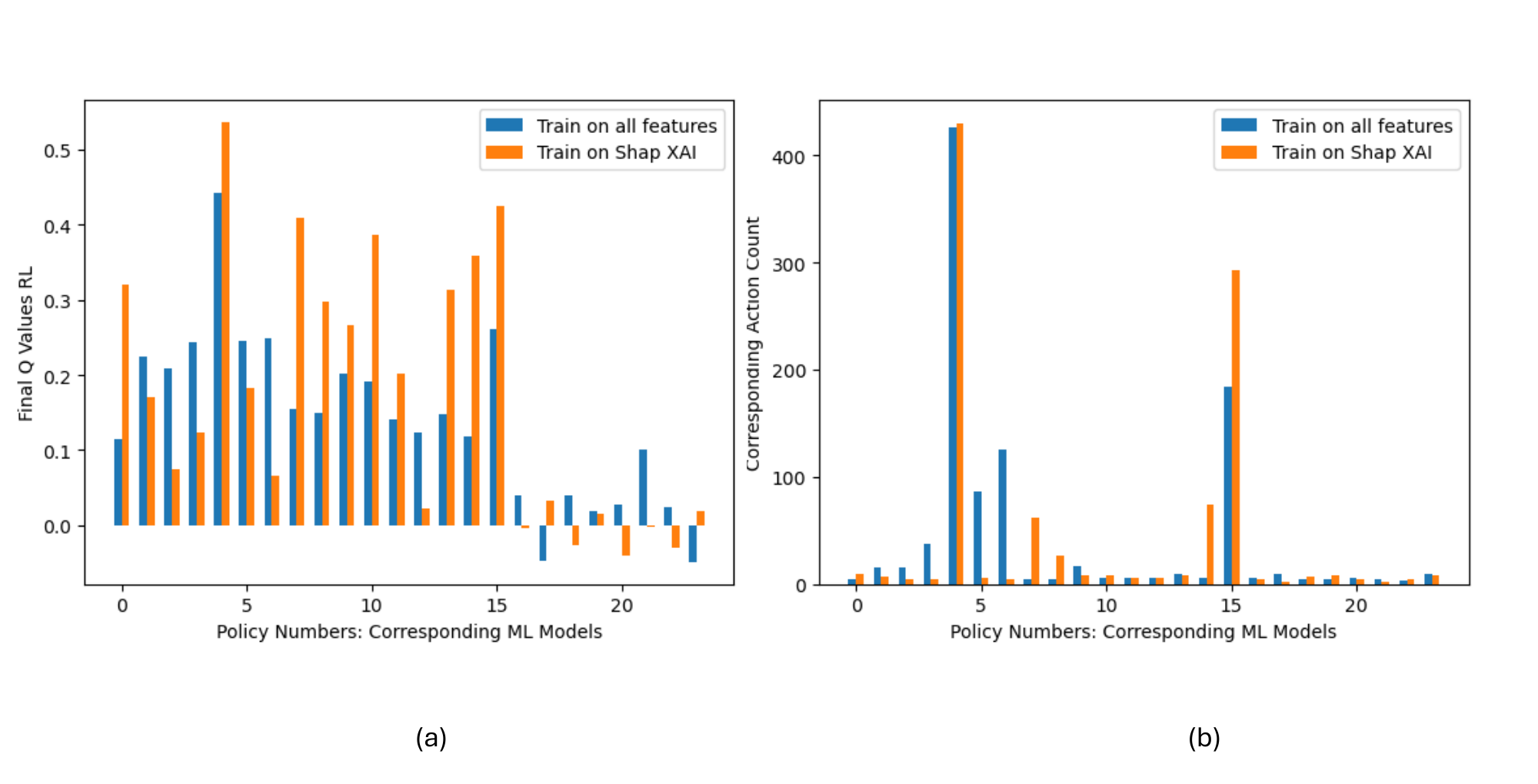}
    \caption{Extracted results from the ADP framework. (a) Bar plot representation of the $Q$ values for each state in the entire search space with two input feature extraction scenarios. (b) Bar plot representing the total number of times the corresponding action of each state is selected by the decision-making unit in the ADP.}
    \label{fig_ADP_bar}
\end{figure*}

\subsection{Analysis of the Extracted Results from the ADP}
Figure \ref{fig_ADP_bar} shows the results from executing the ADP framework. Specifically, Fig. \ref{fig_ADP_bar}(a) illustrates the acquired final $Q$ values corresponding to each of the states within the search space of the higher-level decision-making unit. Figure \ref{fig_ADP_bar}(b) shows the total number based on which the corresponding action of each state is selected by the ADP as the candidate policy in a given episode throughout the ADP framework execution. In addition, the represented information in Fig. \ref{fig_ADP_bar} provides visual comparative analysis highlighting the key differences of the extracted results from the ADP with different input feature selection methods. Namely, the blue bar plots represent the extracted results from the ADP while the corresponding ML models were trained upon all of the input features based on the selected action of each state. However, the orange bar plots illustrate the results of the ADP execution where the top $k$ input features were selected for the ML pipeline of each state through the SHAP XAI analysis. Analysis of the represented results in Fig. \ref{fig_ADP_bar} shows that applying the top optimal input features as a subset of the entire input feature space generally improved the ML model performances. Specifically, while the average value of the accumulated $Q$ values received a score of 0.1895, the average $Q$ value acquired by the states through the entire search space improved to 0.3101 when the ADP was executed using the top $k$ input features extracted from the XAI analysis. However, observations reveal that while the selected optimal top $k$ features improved the performance of some ML models, it had negative effect on the rest of the ML models. This observation reflects the inherent differences in how individual classifiers utilize feature spaces. Certain models depend on the diversity and redundancy of the full feature set to maintain stable decision boundaries, and the removal of weakly informative or correlated variables can reduce robustness in their internal representations, leading to lower reward values during ADP execution. In contrast, other models benefit from feature reduction since the elimination of irrelevant inputs enhances their discriminative capability and mitigates overfitting. As the XAI based feature ranking is derived from the learned behavior of a specific model instance, the resulting top k subset is not universally optimal across all models within the ADP search space, which naturally explains the observed variation in reward outcomes. Hence, input feature selection is not always guaranteed for all of the ML models. In addition, Fig. \ref{fig_ADP_bar}(a) shows that the State 4 achieved the highest $Q$ value in both feature selection scenarios. Thus, State 4 was selected as the optimal state by the ADP in both cases and the corresponding action of State 4 was selected as the optimal policy by the decision-making unit of the ADP. Specifically, State 4 achieved a final $Q$ value of 0.4317 where all of the corresponding input features were utilized for training of the ML model. However, the same state achieved a final $Q$ value of 0.5469 where the top $k$ input features, extracted from the SHAP XAI analysis, were utilized. Similarly, Fig. \ref{fig_ADP_bar}(a), shows that State 15 achieved the second highest $Q$ values in both ADP execution scenarios. Namely, the cumulative final $Q$ value for State 15 was 0.2541 with the initial execution of ADP algorithm. While, benefiting from the SHAP XAI analysis and using only the input features with higher XAI scores excelled the final $Q$ value for State 15 to a final value of 0.4312. Additionally, analysis of Fig. \ref{fig_ADP_bar}(b), indicate that the correlating actions of State 4 and State 15 were selected by the higher-level decision-making unit of the ADP for over 600 times across the entire 1000 iterations. Hence, it could be derived that the ADP framework converged to the optimal solution over 1000 iterations rendering the corresponding action of the State 4 as the optimal policy and the action representation of the State 15, as the second optimal policy. 

\begin{figure*}[!h]
    \centering
    \includegraphics[width=0.8\linewidth]{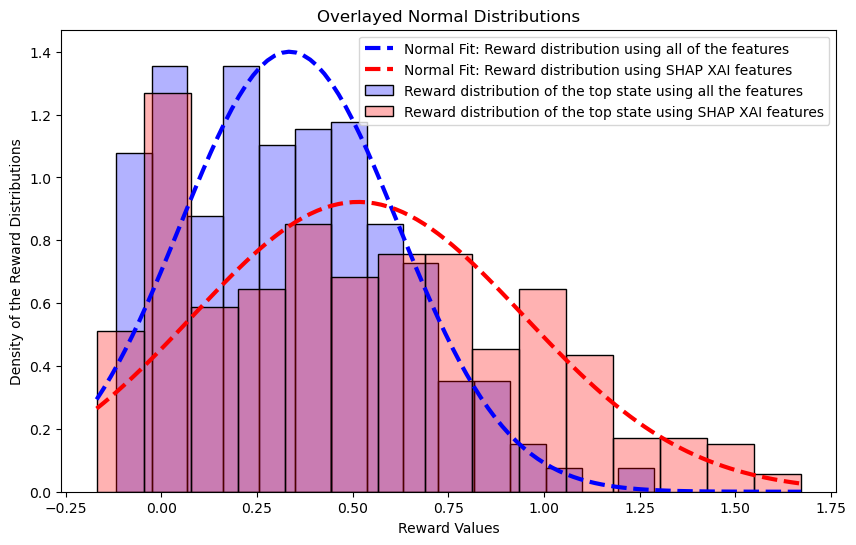}
    \caption{Representation of the Gaussian distribution of the reward values achieved State 4 through ADP framework. The blue histogram plot represents the reward distribution of the optimal state (State 4) during the ADP execution where all of the corresponding input features were utilized to train the ML model. While, the red histogram plot shows the reward distributions of the same optimal state during the ADP execution where the top $k$ input features, derived from the SHAP XAI analysis, were utilized.}
    \label{fig_reward_distribution}
\end{figure*}

Additional analysis were carried out on the reward distributions of the optimal state through ADP executions for better inference of the effectiveness of utilizing SHAP XAI analysis in the output of the ADP framework. Hence, Fig. \ref{fig_reward_distribution} is a representation of the Gaussian distribution of the achieved reward values through extracting the action correlating to the optimal state. In Fig. \ref{fig_reward_distribution}, the blue histogram illustrates the reward distribution for the optimal state (State 4) during the ADP execution when all input features were used to train the ML model. In contrast, the red histogram depicts the reward distribution for the same optimal state, but with only the top $k$ input features, as identified through the SHAP XAI analysis, utilized in the training process. As it is seen in Fig. \ref{fig_reward_distribution}, it could be concluded that using the top $k$ input features, the ML model corresponding to State 4 yielded moderately higher accuracy and $F1$ score values, generally resulting in higher reward values. Specifically, the quantitative analysis of the reward distributions show that while the initial approach achieved an average reward value of 0.3347, the second approach with the SHAP XAI features got a total mean reward value of 0.5148. Table \ref{tab_statistics} shows a comprehensive quantitative analysis of the reward distributions for the top state. Generally, Table \ref{tab_statistics} shows that the ADP strategy using SHAP XAI features outperforms its counterpart by achieving median (0.4857 vs. 0.3138) reward, indicating improved central tendency and overall performance. Additionally, although the standard deviation is slightly higher, the reduced skewness (0.3753 vs. 0.4955) suggests a more balanced and symmetric distribution of rewards, which reflects greater consistency.

For further analysis of the reward distributions, Jensen-Shannon (JS) divergence was conducted on the achieved reward values. The Jensen-Shannon (JS) divergence is a symmetric and smoothed measure of similarity between two probability distributions \(P\) and \(Q\) (\cite{menendez1997jensen}). It is based on the Kullback-Leibler (KL) divergence but has the advantage of always being finite and symmetric. Here $P$ represents the array that stores the reward distribution of the ADP with the first strategy while $Q$ is the array representing the reward distribution of the ADP with SHAP XAI features.

\begin{table*}[]
    \centering
    \caption{Representation of the statistical analysis carried out on the reward value distributions of the top state upon different ADP strategies. The first row shows the quantitative results for the ADP where all of the input features representing the top state were used, while the second row represents the results for the ADP using the SHAP XAI analysis.}
    \begin{tabular}{c|c|c|c|c|c}
    \hline
    \multicolumn{6}{c}{Statistical Analysis of the Acquired Reward Values for the Top State}\\
    \hline
       Row & Strategy  & Mean ($\mu$) & Median ($Q_2$) & \makecell{Standard \\Deviation ($\sigma$)} & Skewness ($\gamma_1$) \\
       \hline
       1 & \makecell{ADP with all\\ the features}  & 0.3347 & 0.3138 & 0.2844 & 0.4955 \\
       2 & \makecell{ADP with \\SHAP features} & 0.5148 & 0.4857 & 0.4325 & 0.3753\\
       \hline
    \end{tabular}
    \label{tab_statistics}
\end{table*}

\begin{table*}[!t]
    \centering
    \caption{Comparison of the top 5 performing states under the ADP framework using (i) all input features and (ii) SHAP-based feature selection.} 
    \begin{tabular}{c|c|c|c|c|c}
    \hline
    & Top states & Final $Q$ & ML & Input feature & top $k$\\
    & & values & Model & strategy ($A_{FE}$) & features \\
    \hline
        \multirow{5}{*}{\makecell{ADP using\\ all features}} &  3 & 0.2445 & \makecell{Random \\Forest} & 4 & All features\\
        & \textbf{4} & \textbf{0.4429} & \makecell{\textbf{Random} \\\textbf{Forest}}  & \textbf{5} & \textbf{All features} \\
        & 5 & 0.2462 & \makecell{Random \\Forest} & 6 & All features\\
        & 6 & 0.2498 & \makecell{Random \\Forest} & 7 & All features\\
        & 15 & 0.2621 & XGboost & 8 & All features\\
        \hline 
        \hline
        \multirow{5}{*}{\makecell{ADP using \\SHAP XAI \\results}} &  \textbf{4} & \textbf{0.5369} & \makecell{\textbf{Random} \\\textbf{Forest}} & \textbf{5} & \textbf{top 20}\\
        & 7 & 0.3753 & \makecell{Random \\Forest} & 8 & top 20 \\
        & 10 & 0.3864 & XGboost & 3 & top 15\\
        & 14 & 0.3597 & XGboost & 7 & top 25\\
        & 15 & 0.4093 & XGboost & 8 & top 20\\
        \hline
    \end{tabular}
    \label{tab_top_5_states}
\end{table*}

\begin{equation}
\text{JS}(P \| Q) = \frac{1}{2} \text{KL}(P \| M) + \frac{1}{2} \text{KL}(Q \| M)
\end{equation}

\vspace{6pt}

Here, \(M\) is the average (or mixture) distribution of \(P\) and \(Q\), defined as follows.

\begin{equation}
M = \frac{1}{2}(P + Q)
\end{equation}

\vspace{6pt}

The KL divergence measures how one probability distribution diverges from a second, expected distribution and is defined as follows.

\begin{equation}
\text{KL}(P \| Q) = \sum_i P(i) \log \frac{P(i)}{Q(i)}
\end{equation}

The JS divergence smooths and symmetrizes the KL divergence by computing the average divergence of each distribution from the mixture \(M\). This makes it more suitable for comparing probability distributions in practical applications, such as machine learning, information theory, and natural language processing. The calculated JS divergence value was 0.0875. The JS divergence value of 0.0875 indicates that the reward distributions of the two ADP strategies are highly similar with only a small degree of divergence. This suggests that while the SHAP-based feature selection alters the reward landscape slightly, it preserves the overall distributional behavior of the original ADP strategy.

\subsection{Evaluation of ADP results}

\begin{figure*}[!h]
    \centering
    \begin{subfigure}[h!]{0.9\linewidth}
        \centering
        \includegraphics[width=\linewidth]{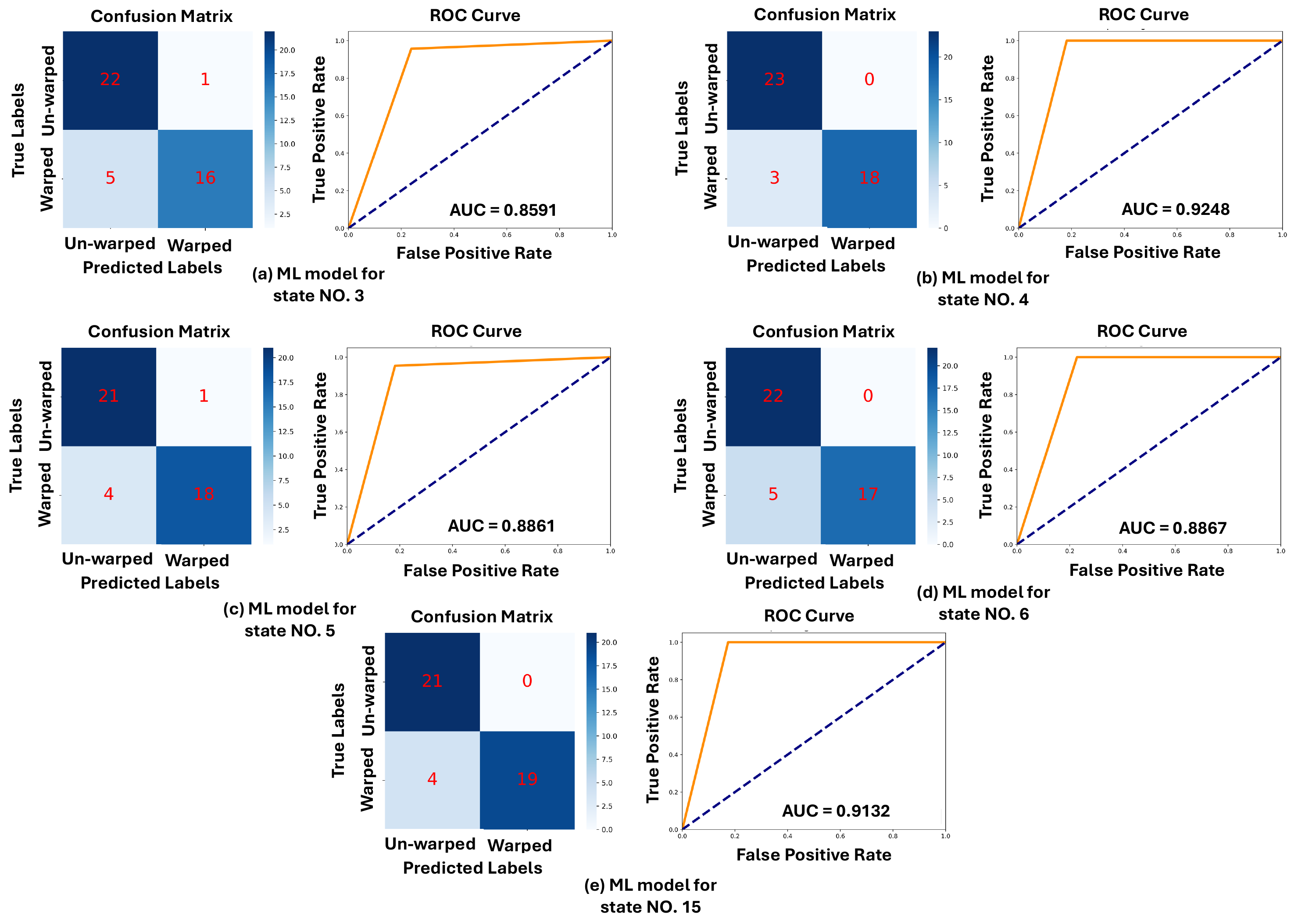}
        \caption{ADP framework using all features}
        \label{fig:adp_all_features}
    \end{subfigure}
    \hfill
    \begin{subfigure}[h!]{0.9\linewidth}
        \centering
        \includegraphics[width=\linewidth]{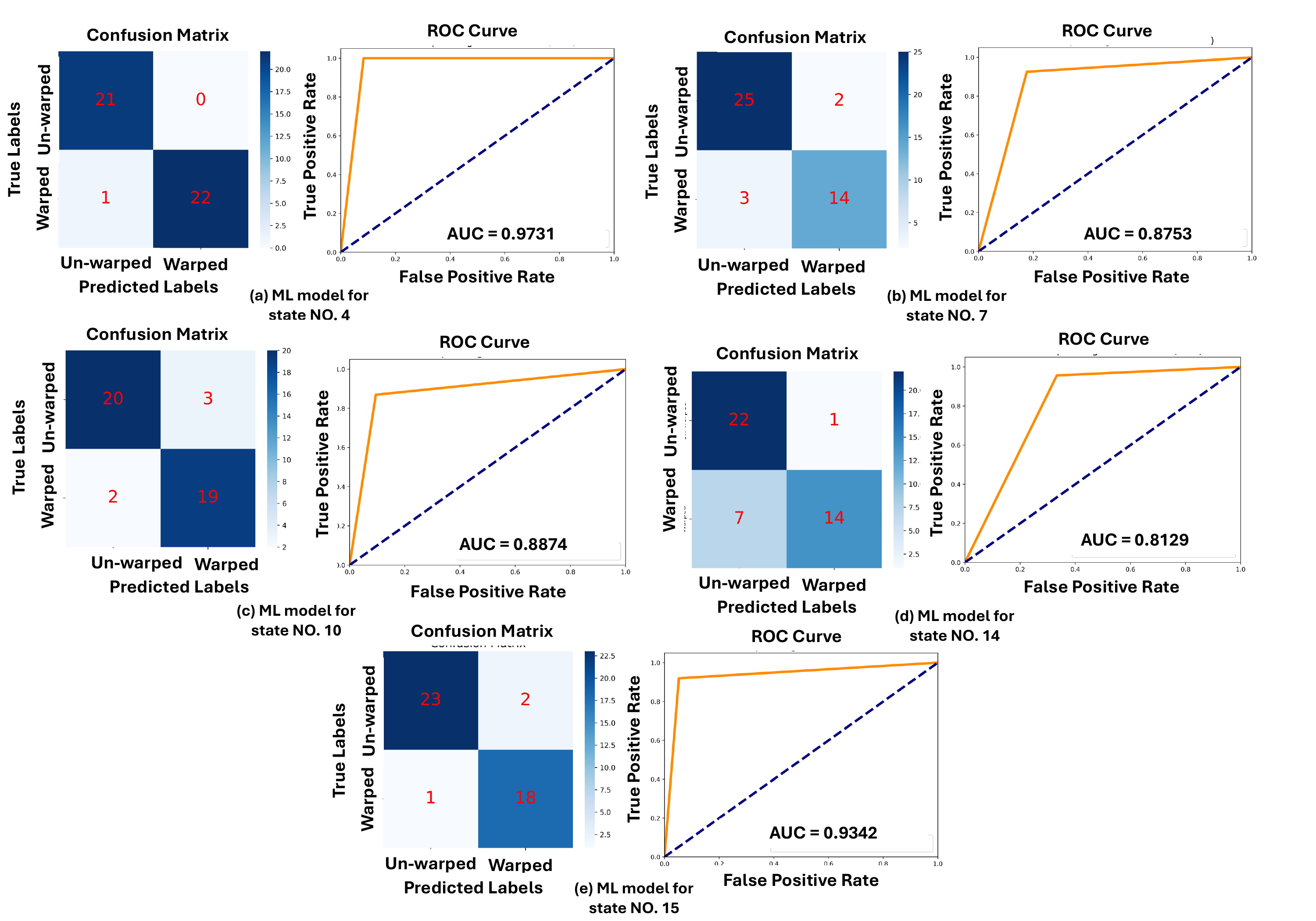}
        \caption{ADP framework using SHAP-selected features}
        \label{fig:adp_shap_features}
    \end{subfigure}
    \caption{Comparison of ADP reward outcomes under two strategies.}
    \label{fig_adp_comparison}
\end{figure*}

Table \ref{tab_top_5_states} shows the corresponding ML models of the top 5 states with the highest $Q$ values extracted from the ADP through both methods. Additionally, as shown in Fig. \ref{fig_adp_comparison}, the ML model associated with State 4 exhibits notably strong classification performance under both ADP strategies. In Fig. \ref{fig_adp_comparison}(a), where all features were used, the model for State 4 achieved an AUC of 0.9248. Although the confusion matrix indicates strong classification performance, minor misclassifications are still present, where three unwarped samples were incorrectly classified as warped, which explains the AUC value being lower than a near-perfect score. All AUC values reported in this study are calculated solely on the held-out test data that remain completely unseen during model training, validation, decision-state evaluation, and reward computation. For every combination of printing parameters, twenty percent of the samples were removed beforehand and used exclusively as the test set. The remaining eighty percent were allocated to training and validation through a five-fold cross-validation scheme within each state of the ADP framework. This strict partitioning ensures that the ADP agent does not access the test data during learning.

If overfitting were present, the models would show high performance on the training and validation folds but substantially lower performance on the unseen test set. This behavior was not observed. The states that achieved high AUC values did so on the test data only, indicating that the models learned reproducible decision boundaries rather than memorizing the training samples. In Fig. \ref{fig_adp_comparison}(b), which uses SHAP-selected features, the AUC for State 4 increased to 0.9731, demonstrating a noticeable improvement in predictive performance. The confusion matrix also shows enhanced classification accuracy compared to Fig. \ref{fig_adp_comparison}(a), however, one unwarped instance was still misclassified as warped. Despite this minor error, the SHAP-based model clearly outperforms the model trained on all features in both discrimination capability and overall robustness. The retention of high accuracy with a reduced feature set demonstrates that the SHAP method effectively eliminates irrelevant information while maintaining model fidelity. This not only confirms the robustness of the ML model for State 4, but also highlights the efficiency gained through explainable AI-driven feature selection. 

Consequently, these results validate the reliability and adaptability of the proposed ADP framework in selecting optimal strategies under different state representations. In comparison, other ML models corresponding to different states show a slight drop in AUC values and a few misclassifications, particularly under the all-features strategy. However, the SHAP-based feature selection consistently enhances or maintains the classification performance across most states, as evidenced by generally improved AUC values and more balanced confusion matrices. This further supports the notion that the SHAP approach enhances model generalization by focusing on the most informative inputs.

\subsection{Alternative Feature Selection Methods}

While SHAP XAI is a powerful method for feature selection, it is not the only available approach. Several other techniques can be employed to reduce input dimensionality and improve model efficiency, including unsupervised and information-theoretic methods such as Principal Component Analysis (PCA) (\cite{abdi2010principal}) and Mutual Information (MI) (\cite{carrara2020estimation}). PCA is a statistical technique that transforms the original correlated features into a new set of uncorrelated variables called principal components, ordered by the amount of variance they capture from the data. This method is especially useful when the goal is to reduce redundancy and retain the most significant data variation without relying on target labels. In contrast, MI is an information-theoretic method that measures the amount of information one random variable shares with another. In the context of feature selection, it quantifies the dependency between an input feature and the target variable. Mathematically, the mutual information \( I(X; Y) \) between two discrete random variables \( X \) and \( Y \) is defined as follows.

\begin{equation}\label{MI_Eq}
I(X; Y) = \sum_{x \in X} \sum_{y \in Y} p(x, y) \log \left( \frac{p(x, y)}{p(x)p(y)} \right)
\end{equation}

\vspace{6pt}

\noindent where \( p(x, y) \) is the joint probability distribution of \( X \) and \( Y \), and \( p(x) \), \( p(y) \) are the marginal distributions. A higher MI value indicates a stronger dependency between the feature and the output, making it a good candidate for selection. 

\begin{figure*}[!t]
    \centering
    \includegraphics[width=0.6\linewidth]{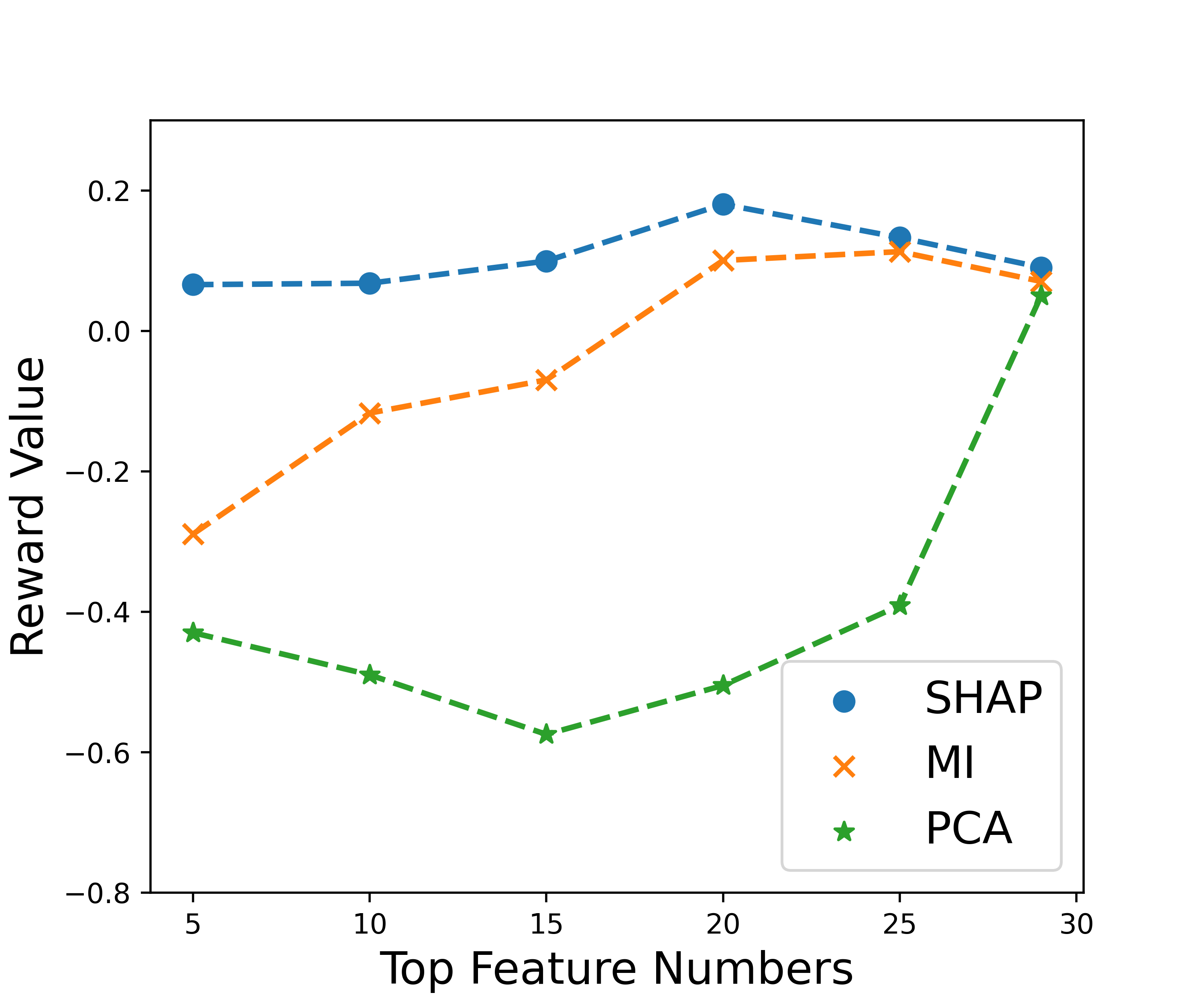}
    \caption{Comparison of the acquired reward value by the ML model corresponding to the top ADP state using the top $k$ features extracted through SHAP XAI, MI, and PCA algorithms.}
    \label{fig_comparison}
\end{figure*}

While SHAP XAI offers model-specific explanations and interpretable feature attribution, PCA and mutual information are alternative methods that can be used for feature selection. For further analysis, both PCA and MI methods were utilized on the ML model of the top state achieved by ADP (State 4). Similar to SHAP XAI, the top $k \in \{5, 10, 15, 20, 25, All\}$ features extracted from the PCA and MI. The model was trained for 10 episodes in each scenario where in the first scenario, the top $k$ features from the SHAP XAI were used, while in the second and the third scenarios, top $k$ features from the MI And the PCA algorithm were used, respectively. The reward value was calculated through extracting the accuracy and $F1$ score on the testing datasets. The extracted results are illustrated as shown in Fig. \ref{fig_comparison}. As Fig. \ref{fig_comparison} shows, the model resulted in better performance in general while using the top features extracted from the SHAP XAI. The results also show a consistency between the output of the developed framework and the manual experiments where using the top $k: k = 20$ SHAP XAI features outputted the highest reward value for the ML model corresponding to State 4. Figure \ref{fig_comparison} also shows that PCA features did not improve the model performance. This observation can be justified based on the linear behavior within the PCA algorithm itself. Namely, the PCA algorithm is an unsupervised linear transformation approach for down-scaling the input feature dimensions. Thus, the most of the nonlinear information within the data is removed by using PCA. Yet, the data collected in this research is representative of a highly nonlinear physical event with significant nonlinear dependency between two respective input features within the input feature space. Additionally, as mentioned previously, PCA is an unsupervised transformation algorithm where it does not include the correlations between the input features within their output counterparts resulting in poor input feature transformations without any consideration of the information that the output data could hold worthy. Previous research has also proved that PCA is not an optimal feature selection method given the datasets within their respective fields (\cite{zhang2024assessing, janecek2008comparison, shi2023research}). 

\begin{figure*}
    \centering
    \includegraphics[width=1\linewidth]{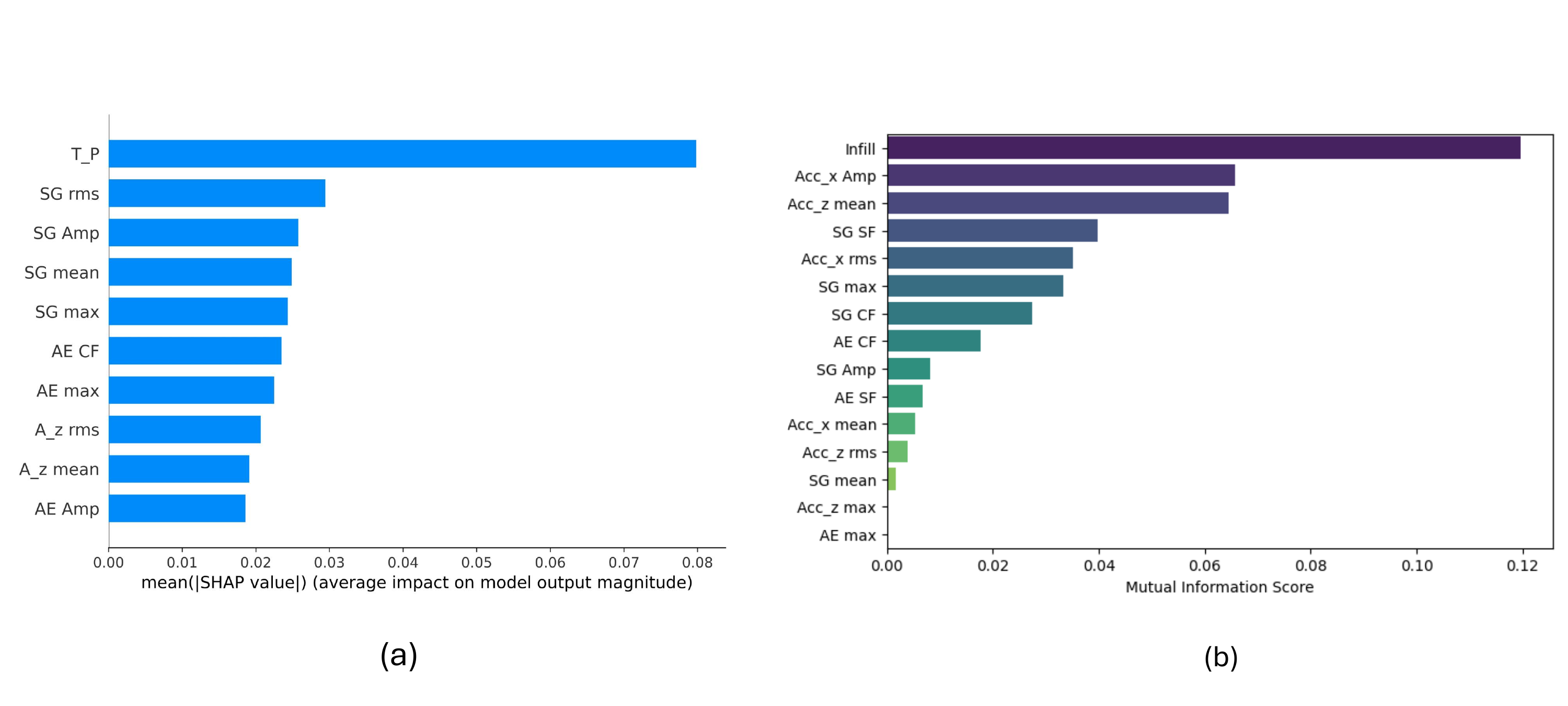}
    \caption{Representation of the top features extracted through the SHAP XAI and the MI algorithms. (a) shows the top 10 features from the SHAP XAI for the ML model representing the top ADP state (State 4) while (b) shows the top 15 features from the MI algorithm of the same ML model.}
    \label{fig_compare_SHAP_MI}
\end{figure*}

Additionally, the results in Fig. \ref{fig_comparison} show that the top features through the MI algorithm have generally resulted in lower ML performance compared to SHAP XAI. This is because how the MI and SHAP XAI interpret the input feature importance. While both SHAP XAI and MI consider the correlations between certain input features and their respective output values, there is a fundamental difference in how this interpretation is extracted. MI relies on the fundamentals of information theory, as represented in Equation \ref{MI_Eq}. Therefore, MI does not consider the ML model performance and how a certain input feature results in the prediction of the ML model. However, SHAP XAI extracts the feature correspondence to the output values significantly based on the performance of the ML model upon which SHAP XAI is applied. As a result, SHAP is more capable of identifying features that are contextually important within the structure of a specific model. This model-awareness enables SHAP to capture nonlinear interactions and dependencies that MI may overlook. Consequently, features ranked highly by SHAP often lead to better generalization and improved predictive accuracy when used for model training or refinement.

\begin{figure*}[!h]
    \centering
    \includegraphics[width=0.6\linewidth]{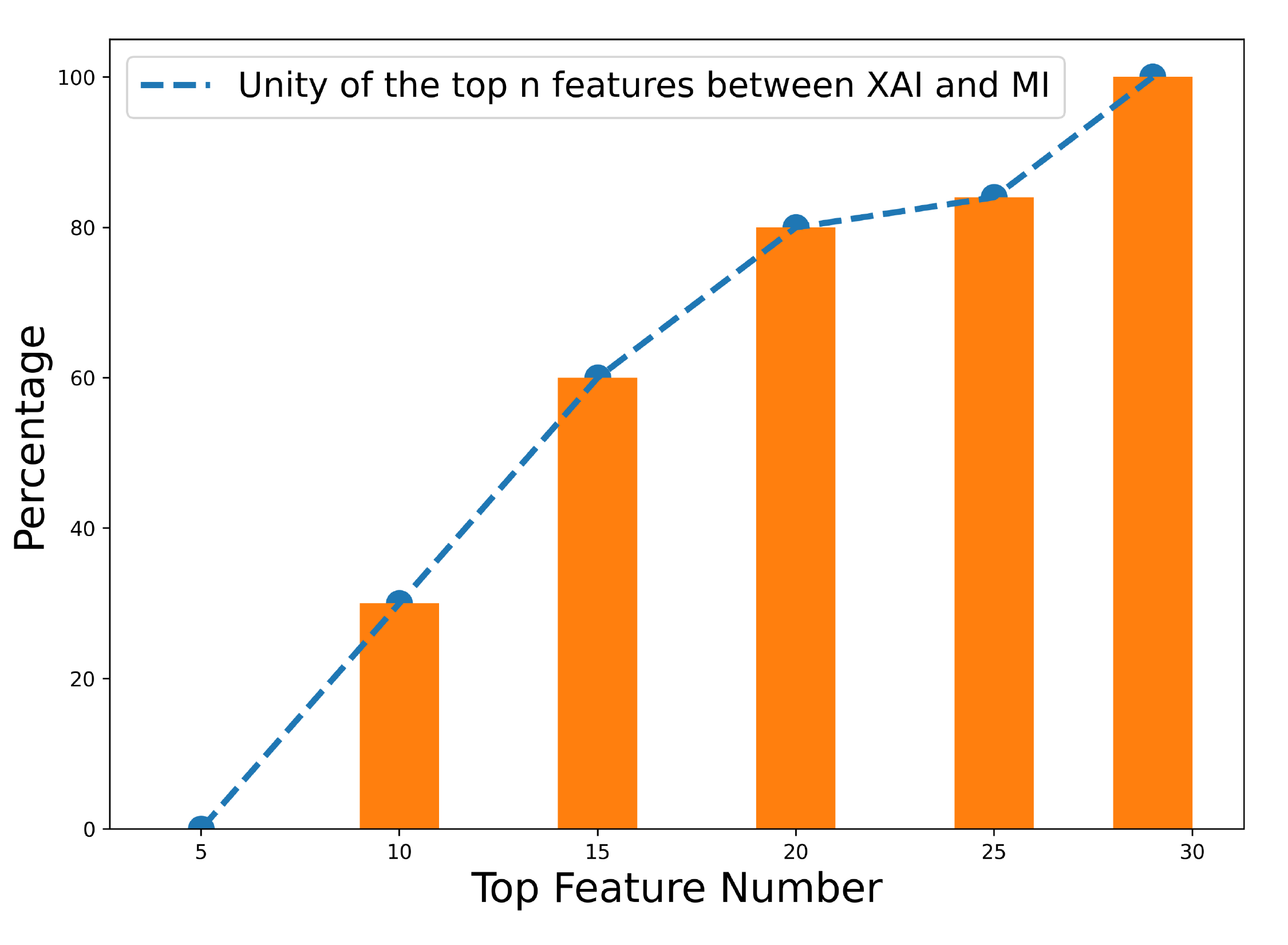}
    \caption{Comparison of feature overlap between SHAP XAI and Mutual Information (MI) as a function of the number of top-ranked features. The orange bars represent the percentage of shared features between the two methods, while the dashed line highlights the increasing agreement as more top features are considered. This visualization underscores the partial alignment between model-aware (SHAP) and model-agnostic (MI) feature ranking techniques.}
    \label{fig_percentage}
\end{figure*}

Further observations in Fig. \ref{fig_comparison} show that the obtained reward value resulted in an overall incremental trend upon increasing the top $k$ features through the MI algorithm. Specifically, the reward value has almost matched the acquired reward value by the SHAP XAI features for top $k$ features where $k \geq 20$. Additionally, Fig. \ref{fig_comparison} indicates that, for the optimal ADP state, the reward achieved using SHAP-selected features consistently surpasses that of MI and PCA across all evaluated $k$ values, with SHAP attaining its highest performance within the range of $k = 20$–$25$ and reaching a maximum reward of 0.2, whereas MI remains comparatively lower and PCA yields negative rewards for the majority of $k$ values. Additional analysis reveals a noticeable difference in the standard deviation of reward values collected over all top-$k$ features, where SHAP-based feature selection exhibits a standard deviation of 0.05, while MI and PCA show higher variability with standard deviations of 0.13 and 0.24, respectively. For further analysis, Fig. \ref{fig_compare_SHAP_MI} is presented to show the top $k$ features extracted from the SHAP XAI and the MI algorithms. Figure \ref{fig_compare_SHAP_MI}(a) highlights the top 10 features ranked by SHAP for the ML model corresponding to the top ADP state (State 4). In contrast, Fig. \ref{fig_compare_SHAP_MI}(b) presents the top 15 features based on their mutual information scores for the same model. These representations offer a comparative view of how model-specific and model-agnostic methods evaluate feature importance. The difference in feature rankings underscores the distinct principles underlying SHAP and MI-based interpretations. The results shown in Fig. \ref{fig_compare_SHAP_MI} show that both the MI and the SHAP XAI differ in the interpretations of the top $k$ features. This can be used as a justification on the huge difference between the acquired reward values of the ML model while using top $k < 20$ features based on SHAP XAI, and the MI. In addition to Fig. \ref{fig_compare_SHAP_MI}, quantitative comparisons shown in Fig. \ref{fig_percentage} provide further insights into the differences of the SHAP XAI and MI features. 

Figure \ref{fig_percentage} shows the percentage to which the SHAP XAI and the MI algorithms share the same top $k$ features. As illustrated, the two feature selection methods do not share any common top features when $k = 5$. Rendering the ML model pose totally different performances using two distinct approaches of feature selection when $k = 5$. Fig. \ref{fig_percentage} further indicates that feature overlap between SHAP and MI is only about 30 percent at $k = 10$ and increases to roughly 80–100 percent only when $k$ approaches 25–30, confirming that at the practically relevant lower $k$ range the two methods select substantially different feature sets. Therefore, the observation in Fig. \ref{fig_comparison} is justifiable where it shows a significant reward value divergence between the SHAP XAI method and the MI method. Conversely, Fig. \ref{fig_percentage} shows that as the number of $k$ increases, the two feature selection methods (SHAP XAI and MI) have more similar features. Therefore, the ML model represents reasonable performance consistency between the SHAP XAI feature selection method, and the MI method as the counterpart for the former method as $k$ increases gradually. Therefore, it could be concluded that since the both SHAP XAI and MI represent more similar top features as $k$ increases, the ML model performance divergency starts to fade as it is shown in Fig. \ref{fig_comparison}.

In general, the analysis from the previously discussed results reveal that optimal feature selection method where a subset of input features are intuitively selected helps improve the performance of the developed ADP framework. Therefore, the conjugation of ADP with the XAI based feature selection results in a more computationally efficient feature selection method which further improves the results extracted from the ADP framework. Additionally, as presented in the results, SHAP XAI outperformed alternative feature selection methods considerably. Specifically, SHAP XAI provides a performance advantage over traditional feature-selection methods because it estimates feature importance using the marginal contribution of each feature to the predictions of a trained model. This allows SHAP XAI to capture nonlinear interactions and cross-feature dependencies, which are common in FDM thermal sensing and process-parameter data. In contrast, PCA performs a linear orthogonal transformation that may obscure physically relevant structure, and filter-based methods evaluate features independently without considering interactions. As a result, PCA and filter-based approaches may overlook features whose relevance emerges only when combined with others. By aligning the importance evaluation directly with the predictive behavior of the model, SHAP identifies a more discriminative and physically meaningful subset of features, which explains the improved performance observed in the ADP framework.
 Additional analysis on the performance of the developed ADP+XAI framework revealed the computational efficiency of the current framework. Namely the efficiency of the proposed framework is more meaningfully characterized through its computational complexity, expressed in terms of algorithmic order $O(n)$, which captures how the framework scales with respect to input size and model configurations, providing a hardware-independent assessment of performance and scalability. A detailed discussion of the computational complexity of the ADP-based decision-making framework has been carried out in the authors' prior work (\cite{valizadeh2025automated}), where it was shown that the ADP framework exhibits a lower computational complexity compared to conventional global search and evolutionary methods, such that global search has a time complexity of $O(S \times S \times E)$ where $S$ denotes the total number of states in the search space and $E$ represents the number of episodes required to obtain a macroscopic statistical characterization of the state-space, whereas the ADP framework operates with a complexity of $O(S \times A \times E)$, where $S$ denotes the number of states, $A$ represents the number of possible actions per state, and $E$ is the number of learning episodes required for convergence, making this reinforcement-learning-driven iterative exploration significantly more efficient for large search spaces by selectively exploiting promising regions rather than exhaustively evaluating all state combinations
. Hence, further investigations in the ADP can help leverage the capabilities of such ADP framework to be utilized by various manufacturers to benefit from their data for in-situ monitoring of their processes. 

\section{Conclusion}\label{sec13}

This study presented an ADP framework designed to evaluate and compare machine learning models using both full and SHAP-selected feature sets across 217 datasets derived from FDM processes. By automating the training, evaluation, and comparison of models in a structured and iterative environment, the framework offers a scalable solution for model selection in data-intensive manufacturing scenarios. The integration of SHAP-based feature selection enabled the framework to assess not only model performance but also the influence of individual input features, thereby enhancing interpretability and guiding the development of more efficient models.

The ADP framework’s ability to identify and reinforce high-performing model-feature combinations stemmed from its iterative evaluation structure, where multiple models were tested across both full and SHAP-selected feature sets. Through this structure, the ADP framework tracked performance metrics including predictive accuracy and $F1$-scores across each fold and episode to learn from repeated exposure to different model configurations. As a result, the framework gradually converged to a combinations that consistently yielded better outcomes, effectively learning which features were most relevant and which ML model was best suited for the task. The use of SHAP XAI features contributed a crucial role in this process by reducing the dimensionality of the input space and retaining only the most important variables to improve both computational efficiency and model interpretability.

The evolution of $Q$ values throughout the episodes reflected how the ADP system updated its internal policy in response to the rewards it observes. This dynamic learning process allowed the framework to adjust its strategy over time, becoming increasingly confident in selecting optimal combinations. Reward distribution plots complement this by revealing the consistency of model performance, with tighter distributions indicating more reliable outputs. Additionally, JS divergence was utilized to measure the difference between policies based on full versus SHAP-selected features. Together, these elements not only quantify the framework’s sensitivity to feature space alterations but also validate its capacity for adaptive learning and robust decision-making in automated model evaluation.

Overall, the proposed ADP framework illustrated the viability of an automated, reinforcement learning-inspired approach to model evaluation and feature assessment in manufacturing analytics where the XAI capabilities were utilized to leverage the optimal solution extracted from the ADP framework. Specifically, by integrating performance metrics with explainable AI techniques like SHAP, the framework proved to provide a meaningful step toward intelligent, self-improving data pipelines supporting both predictive accuracy and interpretability in various smart manufacturing applications. Although the proposed pipeline demonstrated strong performance, its accuracy ultimately depends on the quality and diversity of the available data. Limited sampling of certain printing conditions may introduce a risk of inaccurate predictions. The framework partially mitigates this risk through bounded reward updates and explainability-driven feature selection, but future work will incorporate expanded datasets by extending the current framework to multi-modal smart-manufacturing datasets, where combinations of time-series signals, 2D visual images, and 3D point-cloud data can be integrated within the same ADP-driven AutoML process. Developing a unified ADP–XAI pipeline for multi-modal inputs will allow the decision-making unit to evaluate richer sources of information and select appropriate models across heterogeneous data types. Specifically, future studies will investigate the use of alternative explainability techniques such as saliency maps, Grad-CAM, or gradient-based attribution methods for image-based inputs, enabling the ADP framework to leverage XAI methods tailored to different sensing modalities. In addition, relying on tabular RL methods to implement Markov Decision Processes is considered a vital hurdle as the expansion in the search space, and therefore expansion in the number of states, will require extra computation time and resources rendering the current approach unfeasible. Therefore, future work will focus on RL methods inclined toward DL approaches to effectively update the state-space within the decision making unit of the ADP.\\
Additionally, future work will include evaluating the generalizability of the proposed ADP pipeline under out-of-distribution scenarios, such as leave-one-configuration-out tests based on material type, infill density, or printing temperature. To support these studies, domain-adaptation and unsupervised learning strategies such as  unsupervised transfer learning and self-supervised representation learning will be incorporated to assess the robustness of the framework under previously unseen printing conditions.\\

\noindent\textbf{Acknowledgments} This work was supported in part by the US National Science Foundation under Grant 2322532, in part by the Department of Mechanical and Aerospace of Michigan Technological University. \\

\noindent\textbf{Data Availability} The datasets generated and/or analyzed during the current study are available from the corresponding author on a reasonable request.

\noindent\textbf{Appendix}   
The developed code for the current project could be collected in the following link: \textcolor{blue}{\href{https://github.com/SalehML/Codes-for-ADP_XAI-Framework.git}{Project Code}}.

\backmatter




\section*{Declarations}

\noindent\textbf{Conflict of interest} All authors declare no relevant financial or nonfinancial competing interests.\\

\noindent\textbf{Ethical approval} All authors assure the manuscript is not under consideration for publication and has not been published.\\

\noindent\textbf{Consent to participate} All authors consent to publish this research, and there are no potential issues involved in this research.

\bibliography{sn-bibliography}

\end{document}